\documentclass[10pt,twocolumn,letterpaper]{article}

\usepackage[pagenumbers]{iccv} % To force page numbers, e.g. for an arXiv version

\definecolor{iccvblue}{rgb}{0.21,0.49,0.74}
\usepackage[pagebackref,breaklinks,colorlinks,allcolors=iccvblue]{hyperref}

\usepackage{iccv}
\usepackage{times}
\usepackage{epsfig}
\usepackage{graphicx}
\usepackage{amsmath}
\usepackage{amssymb}
\usepackage{bm}
\usepackage{xcolor}
\usepackage{booktabs}
\usepackage{array} % For custom column types
\usepackage{makecell} % For line breaks within cells
\usepackage{arydshln}  % For dashed lines in tables
\usepackage{xcolor}
\usepackage{pifont}
\usepackage{easyReview}
\usepackage{multirow}
\usepackage[accsupp]{axessibility}  % Improves PDF Readability for those with disabilities

\newcommand{\OURS}{SHeaP}

\definecolor{jiapeng}{rgb}{0.2, 0.4,0.9}

\newcommand{\cmark}{\ding{51}}% Check mark
\newcommand{\xmark}{\ding{55}}% Cross mark

\def\paperID{2158} % *** Enter the Paper ID here
\def\confName{ICCV}
\def\confYear{2025}

\title{\OURS{}: Self-Supervised Head Geometry Predictor Learned via 2D Gaussians}

\author{
Liam Schoneveld\textsuperscript{1} \and
Zhe Chen\textsuperscript{1} \and
Davide Davoli\textsuperscript{2} \and
Jiapeng Tang\textsuperscript{3} \and
Saimon Terazawa\textsuperscript{1} \and
Ko Nishino\textsuperscript{4} \and
Matthias Nießner\textsuperscript{3}
}

\begin{document}

\twocolumn[{
\renewcommand\twocolumn[1][]{#1}%
\maketitle

\vspace{-3em}
\begin{center}
\textsuperscript{1}Woven by Toyota \hfill
\begin{tabular}[t]{@{}c@{}}
\textsuperscript{2}Toyota Motor Europe NV/SA \\[-0.2em]
{\footnotesize \textit{associated partner by contracted service}}
\end{tabular}
\hfill
\textsuperscript{3}Technical University of Munich \hfill
\textsuperscript{4}Kyoto University
\end{center}

\centering\includegraphics[width=\linewidth]{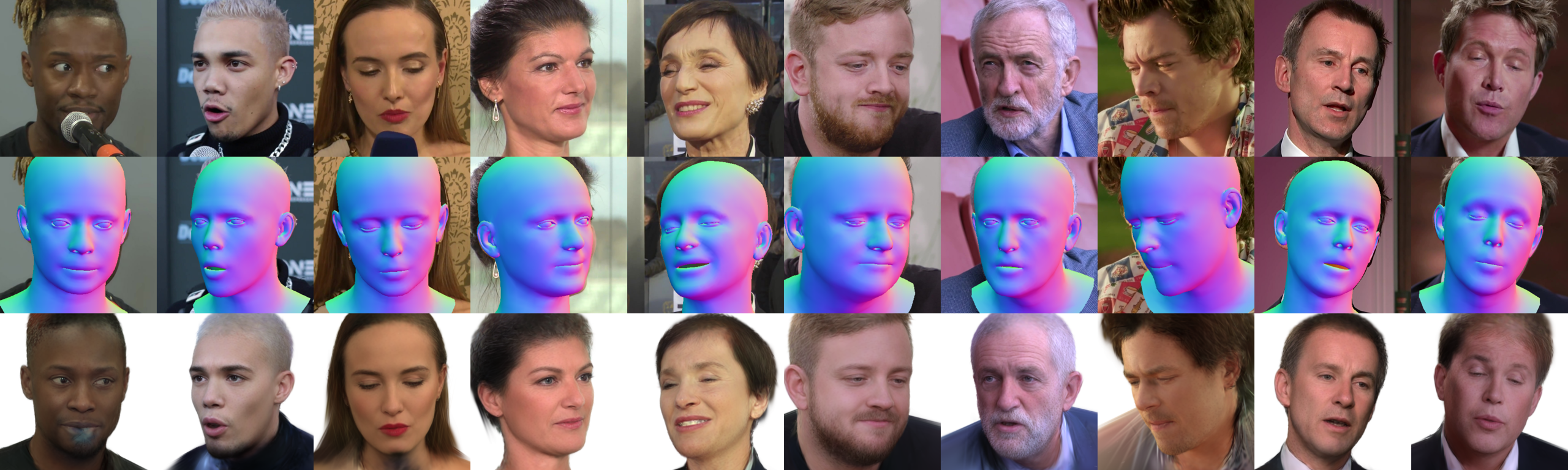}
\captionof{figure}{Our method, named \OURS{}, instantly predicts accurate human head geometry from a single image. First row: input images; second row: predicted meshes; bottom row: rendered predicted Gaussians.}
\label{fig:teaser}
\vspace{2em}
}]
% \MATTHIAS{teaser is not clear as it mostly shows a couple of results; what would be better if it only have of the teaser are results and the rest is along giving a high-level overview of the method; in particular highlight the training process with self-supervision}

\begin{abstract}

Accurate, real-time 3D reconstruction of human heads from monocular images and videos underlies numerous visual applications. As 3D ground truth data is hard to come by at scale, previous methods have sought to learn from abundant 2D videos in a self-supervised manner. Typically, this involves the use of differentiable mesh rendering, which is effective but faces limitations. 
To improve on this, we propose \OURS{} (Self-supervised Head Geometry Predictor Learned via 2D Gaussians).
Given a source image, we predict a 3DMM mesh and a set of Gaussians that are rigged to this mesh. We then reanimate this rigged head avatar to match a target frame, and backpropagate photometric losses to both the 3DMM and Gaussian prediction networks. We find that using Gaussians for rendering substantially improves the effectiveness of this self-supervised approach.
Training solely on 2D data, our method surpasses existing self-supervised approaches in geometric evaluations on the NoW benchmark for neutral faces and a new benchmark for non-neutral expressions. Our method also produces highly expressive meshes, outperforming state-of-the-art in emotion classification.

\end{abstract}

\section{Introduction}

%Motivate task: why is the task important? why specifically do we need a reconstrcuted morphable model? 
% \MATTHIAS{see latex cmments: we are failling to motivate the context for morphable models that can be animated. At the moment it's just a general face reconstruction motivation. }\LIAM{ I updated a bit, let me know if ok now?}\Ko{Needs a killer app for accurate animatable 3D mesh geometry (that's the end result), not photometry (lifelike avatars are mainly that). Could work if you motvate that animatable 3D geometry underlies any photorealistic avatar. If you are starting the next para at the current level, then you will need to motivate the need to do this from an ingle image, here.}

The accurate reconstruction and animation of 3D human head models from single 2D images is a vital task in computer vision, with wide-ranging applications such as lifelike avatars for virtual reality (VR) and augmented reality (AR), realistic digital content creation, and enhanced facial recognition systems. These applications demand precise 3D geometry to achieve high levels of realism and interaction quality. Moreover, many of these applications require real-time processing, necessitating efficient methods that can reconstruct 3D geometry from single 2D images. Estimating accurate 3D head geometry from single images is, however, extremely challenging due to the lack of inherent depth information, further compounded by variations in lighting, expressions, and occlusions. These challenges necessitate innovative approaches that can overcome these limitations while achieving both speed and accuracy.

To simplify the task of human head geometry estimation, 3D Morphable Models (3DMMs) are often used. Offline optimization-based methods have been effective in achieving high-quality reconstructions but are often computationally intensive and unsuitable for real-time applications.
As a result, there has been a shift toward using feedforward neural networks to enable near-instantaneous geometry prediction. Training such networks with direct 3D supervision has been attempted~\cite{zielonka2022mica, tokenface}, but the scarcity of comprehensive 3D datasets limits their scalability and generalization. 
Another line of work has turned to self-supervised learning from 2D data alone~\cite{DECA, EMOCA, deep3dfacerecon, denselandmarks, albedogan, focus, ccface}, using photometric reconstruction losses to guide the learning process. Despite these efforts, current self-supervised approaches struggle with photometric modeling. They often rely on differentiable mesh rendering, which is hampered by the discontinuous nature of mesh rasterization and the lack of realism in the rendered output.

In this work, we introduce a novel approach that enhances self-supervised learning by integrating modern differentiable rendering techniques, specifically Gaussian Splatting~\cite{3DGS, 2DGS}. Our method, Self-Supervised Head Geometry Predictor Learned via 2D Gaussians (\OURS{}), predicts both the 3D Morphable Model (3DMM) parameters~\cite{FLAME} and 2D Gaussians for any given identity from a single image. By employing 3DMM-rigged Gaussian Splatting, we achieve a more detailed and realistic visual representation, allowing for accurate photometric loss computation. Our 2D Gaussians-specific formulation also enhances the coupling between predicted geometry and rendered appearance, which we show is key to the accuracy of the underlying geometry. Lastly, the flexibility of Gaussians allows our approach to also model the hair and shoulders region, obviating the need for meticulous facial masking. 
%Our method demonstrates superior performance on neutral and expressive benchmarks, showcasing the potential of self-supervised approaches trained on scalable 2D data.

Our method outperforms all existing methods trained with only 2D data on the NoW challenge~\cite{NOW}. We also introduce a novel benchmark to evaluate the reconstruction of expressive head geometry using the dense point clouds provided in the Nersemble dataset~\cite{nersemble}. Our method surpasses all publicly available competitors. Finally, our approach achieves state-of-the-art in evaluating the emotional content of predicted 3DMM parameters on AffectNet~\cite{mollahosseini2017affectnet}.

Our contributions can be summarized as follows:

\begin{itemize}
\item We propose a novel approach for learning self-supervised 3DMM predictors by modeling head appearance using 3DMM-rigged Gaussian Splatting. This approach significantly surpasses techniques that rely on differentiable mesh rendering.
% \replace{We show that Gaussian Splatting substantially outperforms differentiable mesh rendering as a visual reconstruction method when used to learn an accurate geometry predictor from only 2D signals.}{}
\item We introduce a novel Gaussians generator architecture that combines a UV-map generator with a graph convolutional neural network. Its design allows for dynamic densification/pruning of Gaussians in the training process, which further improves our method's performance.

\item We present a novel regularization technique that tightly couples the geometry of the 3DMM mesh with that of the Gaussians. Our results demonstrate that this consistency is crucial for achieving precise 3DMM estimation.
% \replace{We introduce new loss terms that leverage specific advantages of 2D Gaussian Splatting, such as accurate normals allowing for additional loss terms and diffuse shading, and show that this is central to our method's success.}{}

% \item \replace{Our method outperforms all existing methods trained with only 2D supervision on the NoW dataset. Furthermore, on a novel benchmark for reconstructing expressive head geometry, our method outperforms all publicly-available competitors. Lastly, our method outperforms the previous state-of-the-art for evaluating the emotion content of predicted 3DMM parameters on AffectNet.}{}

\end{itemize}

\section{Related Work}

%\subsection{3D Face Reconstruction}
3D face reconstruction methods from 2D images/videos can be broadly classified into two categories.

\vspace{-8pt}
\paragraph{Optimization-based methods} estimate face shape and expression by optimizing 3D model parameters to fit 2D observations. Traditional approaches treat this as an inverse rendering problem, leveraging geometric priors from 3DMMs~\cite{blanz2023morphable, 3DMM, FLAME, NPHM, DPHM}, illumination models~\cite{egger2018occlusion}, temporal smoothness constraints, and sparse facial landmark projection consistency for guidance and regularization. To introduce additional constraints, some methods incorporate depth observations~\cite{thies2015real, ImFace, DPHM} and optical flow~\cite{cao2018stabilized} into the optimization process. Recent works have focused on disentangling shape and expression prediction using 3DMMs via analysis-by-synthesis~\cite{thies2016face2face}. Several methods track pose and expression parameters using photometric and sparse facial landmark supervision with differentiable mesh renders~\cite{pytorch3d, laine2020modular}. To enhance supervision, some approaches utilize dense landmarks~\cite{wood20223d} and screen-space UV position maps~\cite{taubner20243d}. However, optimization-based methods generally suffer from slow inference, limiting their practicality in real-time applications.

\vspace{-8pt}
\paragraph{Regression-based methods} employ deep neural networks to reconstruct 3D faces from single images~\cite{DECA, EMOCA, ROME, deep3dfacerecon}. They typically predict 3DMM parameters, camera pose, texture, and lighting conditions. Given the scarcity of 3D-annotated face ground truths, they often rely on self-supervised training on large-scale 2D image datasets using photometric supervision. Most approaches utilize image-classification networks~\cite{he2016deep, sandler2018mobilenetv2} as a backbone for 3DMM parameter prediction. MICA~\cite{zielonka2022mica} was the first to learn metrical face shape priors from 3D-annotated datasets. TokenFace~\cite{tokenface} took this further by combining 3D data with 2D data, utilizing a vision transformer to enhance performance. HRN~\cite{lei2023hierarchical} and SADRNet~\cite{ruan2021sadrnet} adopted a coarse-to-fine hierarchical reconstruction strategy to improve accuracy. EMOCA~\cite{EMOCA} focused more on the emotional content of predicted meshes, improving this through emotion recognition perceptual losses. Lastly, SMIRK~\cite{smirk} also learns to predict emotive meshes, by backpropagating a photometric loss based on reconstructions produced by a neural renderer conditioned on renderings of the predicted 3DMM mesh.

\vspace{-8pt}
\paragraph{Photo-realistic head avatar reconstruction} has seen significant progress with recent neural scene representations such as Neural Radiance Fields (NeRF~\cite{mildenhall2020nerf}), InstantNGP~\cite{mueller2022instant}, and Gaussian Splatting~\cite{3DGS}. It is common for such approaches to integrate a 3DMM, which allows for control of pose and expression, and provides a strong prior. NerFace~\cite{Nerface} introduced expression-dependent NeRFs, controlled by 3DMM coefficients. Nerfies~\cite{park2021nerfies} and HyperNeRF~\cite{park2021hypernerf} introduced time-dependent deformation fields to warp query points from a deformation space to a canonical space, where density and RGB predictions are made. Similarly, INSTA~\cite{zielonka2022insta} and HQ3DAvatar~\cite{HQ3Davatar} employed point-based warping from deformed space back to canonical space, allowing a multi-resolution hash grid to be queried for rendering the head avatar. Despite impressive results, NeRF-based methods suffer from slow training and inference due to the computational cost of dense point sampling and MLP evaluations in volume rendering.

By leveraging explicit rasterization, Gaussian Splatting can render high-resolution images in real time. Although some methods employ Gaussian deformation fields to model facial expression variations~\cite{xiang2024flashavatar, xu2023gaussianheadavatar, giebenhain2024npga}, our work is more closely related to GaussianAvatars~\cite{gaussianavatars, Splattingavatar, tang2024gaf}, which bind splats to the triangles of the FLAME mesh~\cite{FLAME}. These methods propagate pose- and expression-dependent triangular deformations to attached Gaussians. This allows for the Gaussians to be optimized in a canonical space, while the 3DMM parameters are responsible for modeling motion. However, all of these methods assume the 3DMM tracking to be already provided, simplifying the problem to that of just optimizing the Gaussians. We face the far more complex challenge of simultaneously predicting the 3DMM parameters \textit{and} the 3DMM-attached Gaussians.

\section{Preliminaries}

\begin{figure*}[t]
\vspace{-6mm}
  \centering
  \includegraphics[width=1.05\textwidth]{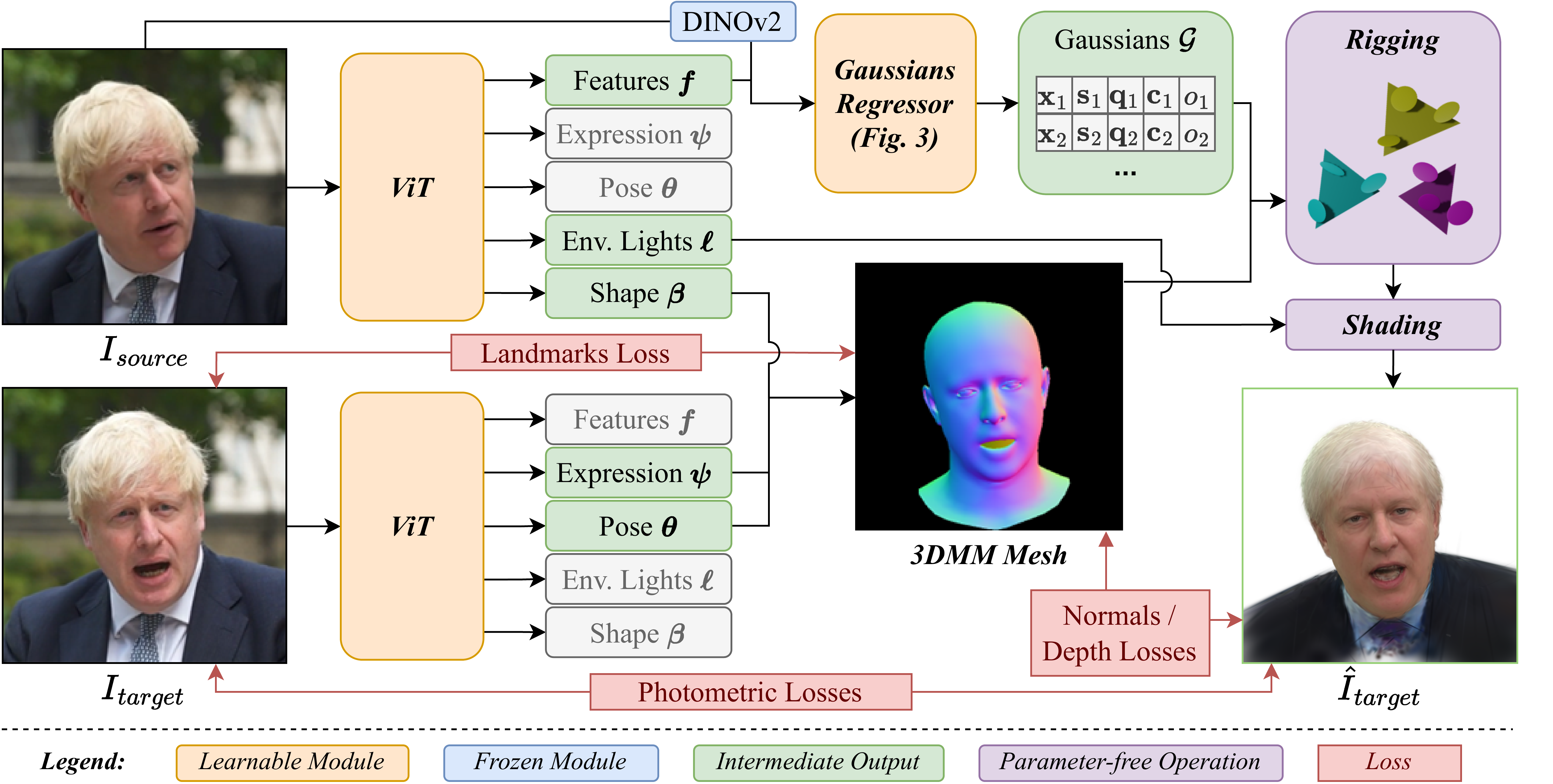}
  \vspace{-6mm}
  \caption{Overview of \OURS{}. At each training step, we sample a source image $I_\textit{source}$ and a target image $I_\textit{target}$. These are both passed through the same vision transformer (ViT), which predicts 3DMM parameters shape $\bm{\beta}$, pose $\bm{\theta}$ and expression $\bm{\psi}$, plus an environment lighting latent $\bm{\ell}$ and identity features $\bm{f}$. A Gaussians Regressor takes $\bm{f}$ as input, along with DINOv2~\cite{dinov2} features $\mathbf{d}$. The Gaussians Regressor predicts a set of Gaussians $\mathcal{G}$, which are bound to the predicted 3DMM mesh and rendered with 2DGS to produce $\hat{I}_\textit{target}$. Finally, photometric losses between $\hat{I}_\textit{target}$ and $I_\textit{target}$ are backpropagated to the ViT and Gaussians Regressor parameters, as well as additional losses based on rendered depth, normals, and landmarks.}
  % (Orange: learnable modules; green: intermediate outputs; purple: operations; red: loss functions.)
  \label{fig:overview}
\end{figure*}

%\subsection{2D and 3D Gaussian Splatting}
3D Gaussian Splatting (3DGS) \cite{3DGS} uses Gaussian primitives to represent 3D scenes, with each primitive defined by a 3D position $\bm{\mu}$ and a 3D covariance matrix $\bm\Sigma$. The covariance matrix $\bm\Sigma$ is decomposed into a rotation matrix $\mathbf{R}$ and a scaling matrix $\mathbf{S}$: $\bm\Sigma = \mathbf{R} \mathbf{S} \mathbf{S}^\top R^\top$. During rendering, these Gaussians are projected onto the image plane. After depth-wise sorting, a tile-based rasterizer performs $\alpha$-blending of all the $N$ primitives that overlap each pixel.

2D Gaussian Splatting (2DGS) \cite{2DGS} removes the third scaling dimension from 3DGS primitives, turning them into 2D surfels that resemble flat discs. 2DGS' ray-splat intersection method to rendering allows for more precise depth maps to be computed, and it also enables closed-form modeling of each splat's normal as the cross-product of its tangents directions. These advantages  enable better geometry estimation.
More specifically, each 2D Gaussian is characterized by its central point, $\bm{\mu} \in \mathbb{R}^3$, and two principal tangent directions, $\mathbf{t_u}, \mathbf{t_v} \in \mathbb{R}^3$, along with their respective scales, $s_u, s_v \in \mathbb{R}$. The normal direction is given by $\mathbf{t_w} = \mathbf{t_u} \times \mathbf{t_v}$. The rotation matrix can be expressed as $\mathbf{R} = [\mathbf{t_u}, \mathbf{t_v}, \mathbf{t_w}]$, while the scaling matrix $\mathbf{S}$ is a diagonal matrix defined as $\text{diag}(s_u, s_v, 0)$. The 2D Gaussian lies on a plane, which can be parameterized in the $uv$-space
\begin{equation}
    P(u,v) = \bm{\mu} + s_u \mathbf{t_u} u + s_v \mathbf{t_v} v = \mathbf{H} [u, v, 1, 1]^\text{T} \,,
\end{equation}
where $\mathbf{H\in \mathbb{R}^{4\times 4}}$ is the plane parameterized with the Gaussian 
\begin{equation}
    \mathbf{H} = \begin{bmatrix}
    s_u \mathbf{t_u} & s_v \mathbf{t_v} & 0 & \bm{\mu} \\
    0 & 0 & 0 & 1
    \end{bmatrix} = \begin{bmatrix}
    \mathbf{R}\mathbf{S} & \bm{\mu} \\
    \mathbf{0} & 1
    \end{bmatrix}\,.
\end{equation}
The opacity of a point in the $uv$-space is obtained by evaluating the Gaussian.%expression:
%\begin{equation}
%    \mathcal{G}(u, v) = \exp\left( - \frac{u^2 + v^2}{2} \right)\,.
%\end{equation}
In practice, these matrices are represented and optimized via a scaling vector $\mathbf{s} \in \mathbb{R}^2$ and a quaternion $\mathbf{q} \in \mathbb{R}^4$. In addition, each Gaussian is represented by an opacity factor $\sigma$ and a color $\mathbf{c}$. In the original implementation~\cite{3DGS, 2DGS}, each Gaussian's color is computed based on the viewing direction via spherical harmonics. These colors are then $\alpha$-blended to give the final pixel colors, as per 3DGS.

% \subsection{GaussianAvatars}

% GaussianAvatars \cite{gaussianavatars} introduced an efficient method for training high quality head avatars from multi-view RGB video. Gaussians are bound to an underlying 3DMM mesh, allowing the avatar to be animated by the 3DMM parameters. In this model, each Gaussian is associated with a single parent triangle in a canonical (local) space. During rendering in a deformed space, each Gaussian undergoes transformations based on its parent's current state, including the relative rotation matrix $R_p$, the isotropic scale determined by the relative area of the triangle $s_p$, and the relative barycenter position of the triangle $T_p$ in world space. The transformations are given by the equations $R = R_p R_c$, $\mu = s_p R_p \mu_c + T_p$, and $S = s_p S_c$, where $R_c$ represents canonical rotations, $\mu_c$ the canonical position, and $S_c$ the canonical scaling of the parent's triangle.

\section{Methodology}

\OURS{} predicts 3DMM parameters for head geometry, and Gaussians for appearance, from a single input image. It is trained only on in-the-wild 2D videos. Training follows a self-supervised paradigm similar to DECA and other works \cite{DECA, ROME, GAGAvatar}, whereby identity information is extracted from a \textit{source} frame and then re-animated with motion information from a \textit{driver} frame. Photometric and landmarks losses between the ground truth and reconstructed target image provide the training signal. Our model consists of two main components (Fig. \ref{fig:overview}): the 3DMM parameter estimator and the Gaussians Regressor (Fig. \ref{fig:GaussiansRegressor}).

% Two frames -- a \textit{source} frame and a \textit{target} frame -- are selected at random from a video of an individual's head. Shape, identity and other constant features like environment lighting are extracted from the source frame, and this is used to construct a GaussianAvatar. This avatar is then rendered with pose and expression information extracted from the target frame. This reconstruction is then compared with the ground truth target frame, and a photometric loss is backpropagated to the model weights.

\subsection{3DMM Parameters Estimator}

The 3DMM parameter estimator employs a transformer architecture similar to TokenFace \cite{tokenface}. The input face image is divided into patches, flattened and added to position embeddings to build tokens. These tokens are then passed to a vision transformer (ViT, \cite{dosovitskiy2020image}), along with five additional learnable tokens representing shape $\bm{\beta}$, expression $\bm{\psi}$, pose $\bm{\theta}$, environment lighting $\bm{\ell}$, and features $\bm{f}$ that are sent to the Gaussians Regressor (see \ref{sec:GaussiansRegressor}). Layer normalization is applied at the output layer, and then a set of final MLP layers produce predictions for the various 3DMM parameter components. We initialize the ViT weights with FaRL \cite{FaRL}.

\subsection{Gaussians Regressor}
\label{sec:GaussiansRegressor}

\begin{figure}[t]
\vspace{-6mm}
  \centering\includegraphics[width=0.48\textwidth]{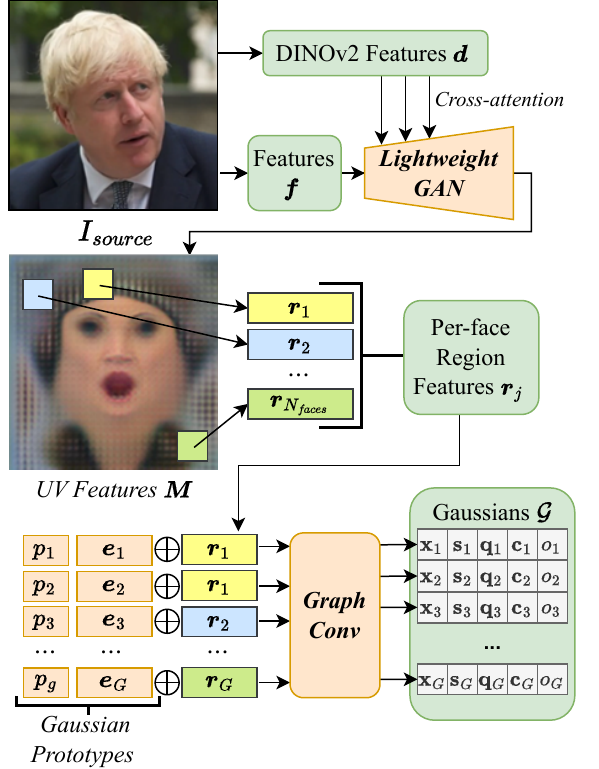}
  \vspace{-6mm}
  \caption{Architecture of the Gaussians generator. In the illustrated case, the first two Gaussians have the same parent face: $p_1 = p_2$ and thus their learned embeddings $\bm{e}_1, \bm{e}_2$  are concatenated with the same region features, $\mathbf{r}_1$.}
  \label{fig:GaussiansRegressor}
  \vspace{-4mm}
\end{figure}

The Gaussians Regressor (Fig. \ref{fig:GaussiansRegressor}) consists of two main networks: a UV map generator that produces UV feature maps $\bm{M}$ given ViT output features $\bm{f}$ and DINOv2 features $\bm{d}$:
$$U: (\bm{f} \in \mathbb{R}^{d_{\text{features}}}, \bm{d} \in \mathbb{R}^{d_{\text{DINO}}}) \to \bm{M} \in \mathbb{R}^{d_{\bm{M}} \times d_{\bm{M}} \times d_{\textit{region}}} \\,$$
and a graph convolutional network that maps Gaussian features $\mathbf{z}$ to actual Gaussians $\mathbf{g}$:
$$\bm{F}: \mathbf{z} \in \mathbb{R}^{N_G \times d_{\mathbf{z}}} \to \mathbf{g} \in \mathbb{R}^{N_G \times d_{\mathbf{g}}} \\.$$
We process the source image through our ViT and take its output $\bm{f}$, and reshape this to a feature map which serves as input to a Lightweight GAN \cite{lightweightgan} architecture. This network cross-attends to DINOv2 features \cite{dinov2}, also extracted from the source image, and produces a $d_{\bm{M}} \times d_{\bm{M}} \times d_{\textit{region}}$ tensor of UV features: $\bm{M}$.

The goal now is to generate a set of $N_G$ Gaussians $\mathcal{G}$, which we will later bind to the 3DMM mesh for rendering. To initialize our set of Gaussians, we assign $N_G^\textit{init}$ Gaussians to each face of the mesh. We refer to the $i$-th Gaussian's parent face index as $p_i$. At initialization, each Gaussian also gets a unique learnable embedding vector $\bm{e}_i$. This is concatenated with `region' features $\bm{r}_i$, which are computed by sampling the generated UV feature map $\bm{M}$:
\begin{equation}
    \bm{r}_i = \textit{GridSample}(\bm{M}, p_i, \bm{V}) \\,
\end{equation}
where the $\textit{GridSample}$ operator here samples the features $\bm{M}$ according to the location of Gaussian $i$'s parent face $p_i$ in the 3DMM's fixed UV map $\bm{V}$.

We concatenate the embedding $\bm{e}_i$ with these region features $\bm{r}_i$ to produce Gaussian features $\mathbf{z}_i$. We want nearby Gaussians to be able to communicate and organize among themselves. To this end, we employ a graph convolutional neural network. This network takes on a ResNet-style architecture \cite{resnet}, with graph-convolution \cite{graphconv} operations replacing convolution operations. The adjacency matrix between Gaussians $i$ and $j$ with parent faces $p_i$ and $p_j$ is defined as:
\begin{equation}
    A_{ij} = 
    \begin{cases}
    1, & \text{if } p_j \in \textit{neighbors}(p_i, \mathcal{D}) \\
    0, & \text{otherwise}
    \end{cases} \\,
\end{equation}
where the $\textit{neighbors}(f, \mathcal{D})$ function finds all neighbors of mesh face $f$ up to degree $\mathcal{D}$ (we set $\mathcal{D}=2$).

The graph convolutional neural network takes as input the adjacency matrix, along with the embeddings and region features, and produces the Gaussians to be rendered. Each Gaussian is defined by 14 values: an offset from the parent face ($\mathbf{x}_i \in \mathbb{R}^3$); a scale ($\mathbf{s}_i \in \mathbb{R}^3$); a rotation quaternion ($\mathbf{q}_i \in \mathbb{R}^4$); an albedo color ($\mathbf{c}_i \in \mathbb{R}^3$); and an opacity scalar ($\sigma_i$).
% \begin{itemize}
%     \item An offset from the parent face $\mathbf{x}_i \in \mathbb{R}^3$;
%     \item A scale $\mathbf{s}_i \in \mathbb{R}^3$;
%     \item A rotation quaternion $\mathbf{q}_i \in \mathbb{R}^4$;
%     \item An albedo color $\mathbf{c}_i \in \mathbb{R}^3$, and;
%     \item An opacity scalar $o_i$.
% \end{itemize}
These Gaussians are then rigged to the 3DMM as described in Section \ref{sec:Binding}.

\subsection{Gaussians Densification and Pruning}

Our novel Gaussians Regressor architecture allows the set of Gaussians $\mathcal{G}$ to undergo a densification and pruning process, like the original 3DGS paper \cite{3DGS}. Rather than densifying and pruning actual Gaussians however, we densify and prune the latent features that produce these Gaussians. 

As introduced in Section \ref{sec:GaussiansRegressor}, each Gaussian is produced by a parent face $p_i$ and an embedding $\bm{e}_i$. Together, we refer to these as Gaussian `prototypes'. For each prototype, we track the opacity values and positional gradients of the Gaussians it produces. Every $t_\textit{densify}$ iterations, we take the $n_\textit{prune}$ prototypes with the lowest average predicted opacity values and delete them from $\mathcal{G}$. We then make a copy of the prototypes that received the largest average positional gradients during the last $t_\textit{history}$ iterations. We then add a small amount of noise to the embeddings of the copied prototypes $\bm{e}_i$, and add them to $\mathcal{G}$. We typically set $n_\textit{prune} = n_\textit{densify}$, which keeps the total number of Gaussians constant. We also enforce that each face on the mesh has at least one Gaussian attached to it, and at most six.

\subsection{Binding Gaussians to 3DMM and Shading}\label{sec:Binding}

\paragraph{Binding.} We build upon the GaussianAvatars~\cite{gaussianavatars} binding formulation by adapting it to 2DGS~\cite{2DGS}. In GaussianAvatars, Gaussians are anchored to an underlying 3DMM mesh, enabling the avatar to be animated using 3DMM parameters. Each Gaussian is linked to a specific parent triangle in a canonical (local) space. During rendering in deformed space, each Gaussian undergoes transformations based on its parent triangle's current state, including the relative rotation matrix $\mathbf{R}_p$, the isotropic scale determined by the relative area of the triangle $s_p$, and the centroid of the triangle $\bm{\mu}_p$ in world space.
% In contrast to GaussianAvatars~\cite{gaussianavatars}, our method necessitates a precise correspondence between each Gaussian and its parent triangle. This precision is essential for accurately aligning the 3DMM geometry with the appearance data derived from the Gaussians. 
% To improve the coupling between Gaussians and their parent faces, we refine the parent triangle scaling of GaussianAvatars' to instead use an anisotropic scale vector \(\mathbf{s_p} =[s_p^u, s_p^v, s_p^n]\). Here, \(s_p^u\) and \(s_p^v\) represent the length of the triangle in the UV directions defined by the matrix $\mathbf{R_p}$. The scale in the normal direction of the triangle \(s_p^n\) is defined as the minimum between \(s_p^u\) and \(s_p^v\). The scaling matrix of the parent triangle is defined as $\mathbf{S_p} = \text{diag}(s_{p, u}, s_{p, v}, 0)$.

To improve the coupling between Gaussians and their parent faces, we refine the parent triangle scaling of GaussianAvatars' to instead use an anisotropic scaling matrix $\mathbf{S}_p = \text{diag}(s_p^u, s_p^v, s_p^n)$. Here, \(s_p^u \in \mathbb{R}\) and \(s_p^v \in \mathbb{R}\) represent the length of the triangle in the UV directions defined by the matrix $\mathbf{R}_p$. The scale in the normal direction of the triangle \(s_p^n \in \mathbb{R}\) is defined as the minimum between \(s_p^u \) and \(s_p^v\). The transformations are given by the equations 
\begin{equation}
  \mathbf{R} = \mathbf{R}_p \mathbf{R}_c \quad \bm\mu = \mathbf{R}_p \mathbf{S}_p \bm\mu_c + \bm{\mu}_p \quad \mathbf{S} = \mathbf{S}_p \mathbf{S_c}  
\end{equation}
where $\mathbf{R_c}$ represents canonical rotations, $\bm{\mu_c}$ the canonical position, and $\mathbf{S_c}$ the canonical scaling of the parent's triangle. 
% Note that while the scaling matrix $\mathbf{S_p}$, with a null scale in the normal direction, affects the final scale $\mathbf{S}$, the scaling vector $\mathbf{s_p}$, which affects the Gaussian's position $\bm{\mu}$, has a non-null normal scale.
Note that the scale in the normal direction, \( s_p^n \), does not influence the final scale of the 2D Gaussian \( \mathbf{S} \), which is defined solely in the UV directions.
However, it does affect the Gaussian’s center position \( \bm{\mu} \), introducing offsets along the surface normal of the mesh. This allows the Gaussian to extend beyond the mesh surface, capturing regions and details not represented by the underlying 3DMM.
% Given a triangle defined by its vertices \(v_1, v_2, v_3\), we first define the triangle rotation matrix \(\mathbf{R_p} = [\mathbf{t_{p,u}}, \mathbf{t_{p,v}}, \mathbf{t_{p,n}}]\) as follows:
% \[
% \mathbf{t_{p,u}} = \frac{v_1 - v_0}{\|v_1 - v_0\|} \quad \mathbf{t_{p,n}} = \mathbf{t_{p,u}} \times \frac{(v_2 - v_0)}{\|v_2 - v_0\|} \quad \mathbf{t_{p,v}} = \mathbf{t_{p,u}} \times \mathbf{t_{p,n}}
% \]
% The components of the scale vector are then defined as:
% \[
% s_{p,u} = \|v_2 - v_1\| \quad s_{p,v} = |\mathbf{t_{p,v}} \cdot (v_2 - v_0)| \quad s_{p,n} = \min(s_{p,u}, s_{p,v})
% \]

% Finally, the homogeneous transformation matrix of the 2D Gaussian is revised as:
% \[
% \mathbf{H} = \begin{bmatrix}
% \mathbf{R_p}\mathbf{R_c}\mathbf{S_p}\mathbf{S_c} & s_{p,v}s_v\mathbf{t_{p,v}}\mathbf{t_v} & 0 & \mathbf{R_p}\mathbf{s_p}\bm{\mu} \\
% 0 & 0 & 0 & 1
% \end{bmatrix}
% \]
% where \([\mathbf{t_u}, \mathbf{t_v}]\) are the tangent directions of the Gaussian in the canonical space, \([s_u, s_v]\) are the scales along the tangent directions, and \(\bm{\mu}\) is the center position of the Gaussian.

\paragraph{Illumination Model.} For Gaussian $i$, the Gaussians Regressor outputs an RGB albedo color $\mathbf{a}_i \in \mathbb{R}^{3}$. To compute the final color of each splat, we combine this with a Lambertian shading model based on Spherical Harmonics (SH) \cite{sphericalharmonics}. The ViT predicts a vector of lighting principal component weights from the source image: $\bm{\ell}_\textit{source}$. These weights are then transformed into spherical harmonics coefficients $\bm{w}_{SH}$ via the Basel Illumination Prior \cite{egger2018occlusion} PCA space $\bm{w}^{SH} = \bm{P} \bm{\ell}_\textit{source}$, where $\bm{P}$ is a $D \times 27$ matrix of eigenvectors. Splitting these weights by color channel $j$, we compute the final color 
\begin{equation}
    c_{ij} = a_{ij} \sum_{k=1}^{9} w_{jk}^{SH} SH_k (N_i) \,,
\end{equation}
where $\mathcal{N}_i$ is the normal of the $i$'th 2D Gaussian (computed as the cross product of its tangent vectors), and $SH_k:  \mathbb{R}^3 \to \mathbb{R} $ defines the SH basis and coefficients.

\subsection{2D Self-Supervised Loss Functions}

After predicting our 3DMM, Gaussians, and rendered image, we can compute a number of self-supervised loss functions by comparing aspects of the rendered images to their ground truth. Unless otherwise stipulated, these losses pertain to the target image and not the source image.

\subsubsection{Landmarks Loss}

We calculate an L1 loss between the projected landmarks of our predicted mesh and 2D landmarks from third party models (\cite{facealignment}, \cite{mediapipe}). We do not perform occlusion or head pose filtering on these landmarks; we just employ a very small loss weight on the landmarks loss, which prevents inaccuracies from having too large of an effect on the model. 

\subsubsection{Photometric Losses}

We compute four photometric losses between $I_\text{target}$ and $\hat{I}_\text{target}$, and denote these together as $\mathcal{L}_\text{photo}$:
\begin{equation}
	\mathcal{L}_\textit{photo} = w_\text{L1} \mathcal{L}_\textit{L1} + w_\text{perc} \mathcal{L}_\textit{perc} + w_\text{ID} \mathcal{L}_\textit{ID} + w_\text{emo} \mathcal{L}_\textit{exp} \,,
\end{equation}
Before computing any of these losses, the background is removed from the target image (using \cite{stylematte}) and replaced with the same random color used as the background when rendering the Gaussians.

\textbf{L1 Photometric Loss} is computed as:
\begin{equation}
	\mathcal{L}_\textit{L1} = \lvert (I_\text{target} - \hat{I}_\text{target}) (0.7 M_\text{face} + 0.3)  \rvert \,,
\end{equation}
where $M_\text{face}$ is a per-pixel mask (predicted with \cite{easyportrait}) which equals $1$ in the face region and $0$ elsewhere.

\textbf{Perceptual Loss.} Here we employ the same facial perceptual loss as \cite{GAGAvatar}. The loss is computed as:
\begin{equation}
	\mathcal{L}_\textit{perc} = \lvert P(I_\text{target}) - P(\hat{I}_\text{target}) \rvert\,,
\end{equation}
where $P$ computes features at various layers from a VGGFace network \cite{vggface}.

\textbf{Identity Divergence Loss} is computed as the cosine similarity of Arcface features computed on the facial region:
\begin{equation}
	\mathcal{L}_\textit{ID} = 1 - \frac{
		f_\text{ID}(I_\text{target}) f_\text{ID}(\hat{I}_\text{target})
	}{
		\lVert f_\text{ID}(I_\text{target}) \cdot f_\text{ID}(\hat{I}_\text{target}) \rVert_2
	}\,.
\end{equation}

\textbf{Facial Expression Loss} is computed as per the identity divergence loss, with features extracted instead from a facial expression recognition network \cite{liamfacialexppaper} $f_\text{exp}$:
\begin{equation}
	\mathcal{L}_\textit{exp} = 1 - \frac{
		f_\text{exp}(I_\text{target}) f_\text{exp}(\hat{I}_\text{target})
	}{
		\lVert f_\text{exp}(I_\text{target}) \cdot f_\text{exp}(\hat{I}_\text{target}) \rVert_2
	}\,.
\end{equation}

\subsubsection{Gaussians vs. 3DMM Geometric Coupling Loss}

We wish to encourage a tight coupling between the 3DMM mesh's geometry and that implied by the Gaussians. To achieve this, we introduce regularization on the rendered normals and depth maps of the 3DMM and Gaussians.

The normal of a given 2D Gaussian is computed as the cross-product of its two tangent vectors. 2DGS rendering thus can compute a normal image by alpha-blending the normals of all primitive along a pixel's ray. We penalize the L1 distance between the rendered 3DMM mesh normals $N_\text{3DMM}$, and the rendered Gaussian normals $N_\text{2DGS}$ (masking out non-facial regions):
\begin{equation}
	\mathcal{L}_\textit{normals} = \lvert
		(N_\text{2DGS} - N_\text{3DMM}) \cdot M_\text{face}
	\rvert \,,.
\end{equation}
To further enforce this geometric similarity between the Gaussians and the 3DMM mesh, we also render depth and compute another L1 loss:
\begin{equation}
	\mathcal{L}_\textit{depth} = \lvert
	(D_\text{2DGS} - D_\text{3DMM}) \cdot M_\text{face}
	\rvert \,,
\end{equation}
where $D_\text{2DGS}$ is the depth rendered from the 2D Gaussians representation.

\subsubsection{Regularizers}

\textbf{Gaussian Regularizers}. We penalise the L2 norm of the offset $\bm{x}_i$ and scale $\bm{s}_i$ of the predicted Gaussians. As per \cite{gghead}, to encourage Gaussians to be either fully transparent or opaque, we also regularize opacity. The total Gaussians regularization loss is:
\begin{equation}
	\mathcal{L}_\textit{reg}^\textit{Gau} = \sum_{i=0}^{N_G} \Big (
	  w_{\bm{x}} \lVert \bm{x}_i \rVert_2 + 
	  w_{\bm{s}} \lVert \bm{s}_i \rVert_2 +
	  w_o \textit{Beta} (o_i)
	  \Big )
	  \,,
\end{equation}
where $\textit{Beta}$ is the negative log-likelihood of a Beta(0.5, 0.5) distribution.

\textbf{3DMM Regularizers.} We regularize the L2 norm of the predicted 3DMM shape and expression codes:

\begin{equation}
	\mathcal{L}_\textit{reg}^\textit{3DMM} = w_{\bm{\psi}} \cdot \lVert \bm{\psi} \rVert_2 + w_{\bm{\beta}} \cdot \lVert \bm{\beta} \rVert_2\,.
\end{equation}

\section{Experimental Setup}

\paragraph{Training Datasets.} We train with VFHQ \cite{VFHQ} and Nersemble \cite{nersemble}. Although Nersemble is a multi-view dataset, we do not make use of its camera extrinsics data, essentially treating it as 2D data. At each training step, we sample a random image to be the source image, and then another random image from the same video to be the target image. We sample from VFHQ~\cite{VFHQ} 75\% of the time and Nersemble~\cite{nersemble} 25\% of the time.

\vspace{-8pt}
\paragraph{Evaluation Methods.}
\label{sec:Evaluation}
We use two datasets for geometric evaluation; both of which require a 3D mesh to be predicted given a single image of a person's face. The predicted 3D mesh is then rigidly aligned to a ground truth scan, and the reported metric is the median and mean distance in millimeters from each point in that scan to a large number of random points sampled from the surface of the mesh. If a scaling factor is fit during this rigid alignment process, then a `non-metrical` evaluation is being performed; otherwise, the evaluation assumes the predicted mesh is already to scale and thus it is a `metrical' evaluation. Here we focus on non-metrical evaluation. 

We evaluate neutral geometry on the NoW dataset~\cite{NOW}. To evaluate non-neutral expressions, we construct a new evaluation metric based on metric-accurate point clouds extracted from Nersemble~\cite{nersemble}. In Nersemble, dense point clouds estimated with COLMAP~\cite{colmap_mvs} are provided for several multi-view frames for 10 subjects. We filter these point clouds to only contain points in the face region, and exclude frames with low confidence, resulting in 60 high-quality point clouds with 960 images captured from different views. For each of these timesteps, we retrieve the multi-view images for which a face is visible, and predict the head mesh geometry with all comparison methods. Like in NoW~\cite{NOW}, we evaluate each method by rigidly aligning the predicted meshes with the ground truth point clouds. We report the median and mean L2 distance in mm from each point in the cloud to its nearest point on the mesh surface.

Finally, we measure the emotional content of the meshes predicted by \OURS{}, by evaluating on AffectNet~\cite{affectnet}. We follow EMOCA's~\cite{EMOCA} approach and first predict 3DMM parameters for all images in the AffectNet dataset, and then fit a 4-layer MLP to predict valence, arousal, and emotion category, based only on these parameters. Our quantitative results are reported in Section \ref{sec:QuantResults}. Further implementation details are provided in the supplementary material.

\begin{figure}[t]
	\centering
	\includegraphics[width=0.48\textwidth]{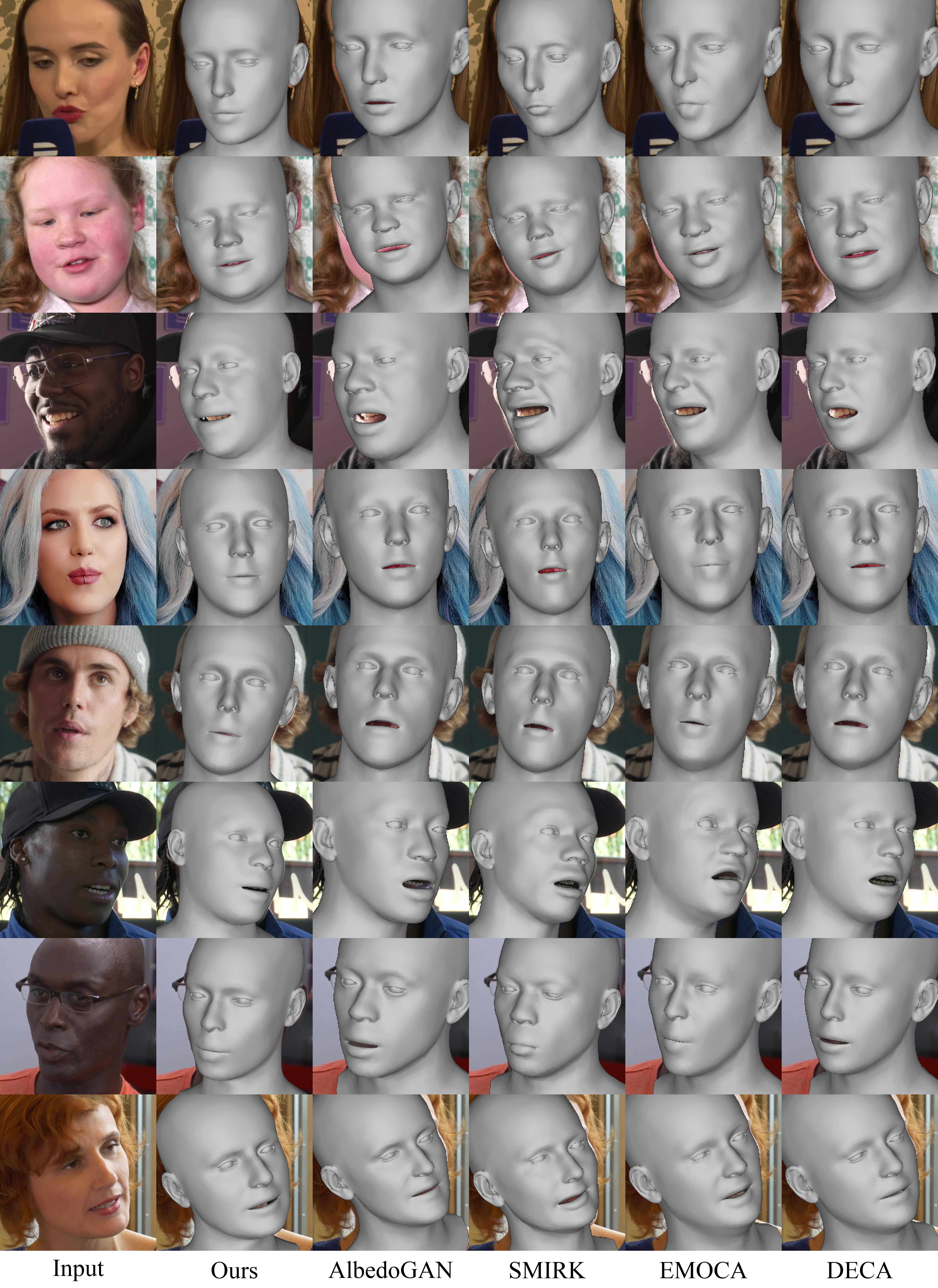}
    \vspace{-6mm}
	\caption{Comparison to other one-shot reconstruction methods, from left to right: Ours, AlbedoGAN \cite{albedogan}, SMIRK \cite{smirk}, EMOCA \cite{EMOCA}, DECA \cite{DECA}.}
	\label{fig:qualitative}
    \vspace{-2mm}
\end{figure}

\vspace{-8pt}
\paragraph{Implementation Details.}
Our method is implemented in PyTorch \cite{pytorch} and uses Pytorch3D \cite{pytorch3d} and gsplat \cite{gsplat}. We use the Adam optimizer \cite{adam}, with a learning rate of 1e-3 for everything except the ViT, which uses 1e-5. Images are compared and rendered at 512x512 resolution, but are resized to 224x224 for the ViT and 448x448 for DINOv2 feature extraction. Preprocessing steps involve cropping to the facial region, and predicting facial landmarks \cite{mediapipe, facealignment}, face segmentation maps \cite{easyportrait}, and background segmentation maps \cite{stylematte}. For rendering of both Gaussians and meshes, we fix the FOV degrees to 14.3, which is the median FOV degrees of Nersemble data downsampled by a factor of 4 and center-cropped to 512x512. Additional details and results are provided in the supplementary materials.

\section{Results}
\begin{table}[t]
\centering
% \setlength{\tabcolsep}{8pt} % Adjust this value as needed
% \begin{tabular}{@{}m{3.5cm}|c|cc@{}}
\setlength{\tabcolsep}{6pt} 
\begin{tabular}{*{4}{c}}
\toprule
 \multirow{2}*{Method} & Only 2D & \multicolumn{2}{c}{Distance (mm)} \\
    \cmidrule(lr){3-4}
    & Supervision & Median & Mean \\ 
\midrule
Deep3DFaceRecon \cite{deep3dfacerecon} & \cmark & 1.11 & 1.4 \\
DECA \cite{DECA} & \cmark & 1.09 & 1.38 \\
CCFace \cite{ccface} & \cmark & 1.08 & 1.35  \\
FOCUS \cite{focus} & \cmark & 1.04 & 1.30 \\
DenseLandmark \cite{denselandmarks} & \cmark & 1.02 & 1.28 \\
AlbedoGAN \cite{albedogan} & \cmark & 0.98 & 1.21 \\
\textbf{Ours} & \cmark & \textbf{0.95} & \textbf{1.18} \\
% \addlinespace % Adds extra space after the dashed line
\cdashline{1-4} % Dashed line from column 1 to column 4
\addlinespace % Adds extra space after the dashed line
MICA~\cite{zielonka2022mica} & \xmark & 0.90 & 1.11 \\
FlowFace~\cite{taubner2024flowface} & \xmark & 0.87 & 1.07 \\
TokenFace~\cite{tokenface} & \xmark & 0.78 & 0.95 \\
\bottomrule
\end{tabular}
\vspace{-2mm}
\caption{Non-metrical evaluation on the NoW Test Set. Our method achieves state-of-the-art error rates among methods that rely only on 2D supervision. Methods below the dashed line use 3D supervision.}
\label{tab:nowresults}
\vspace{-6mm}
\end{table}

\subsection{Qualitative Evaluation}

\OURS{} reconstructs 3D head geometry, pose and expression given a single face image. Fig. \ref{fig:qualitative} qualitatively compares \OURS{} with other state-of-the-art methods, namely Deep 3D Face Reconstruction (Pytorch implementation, \cite{deep3dfacerecon}), DECA \cite{DECA}, EMOCA \cite{EMOCA} and SMIRK \cite{smirk}. \OURS{} produces more plausible geometry that better aligns with the input image. Our method is also the only method capable of disentangling neck from torso pose; all other methods keep the neck joint fixed, rotating the entire torso to align the facial region. This enhances the reconstruction realism and the the overall alignment of the predicted mesh. \OURS{} also predicts realistic and accurate facial expressions, which are focused on accurate geometry reconstruction rather than producing exaggerated expressions. This is key to \OURS{}'s quantitative performance, described in the next section.

\subsection{Quantitative Evaluation}\label{sec:QuantResults}

We compare \OURS{} with other publicly-available models, namely DECA \cite{DECA}, EMOCA \cite{EMOCA}, SMIRK \cite{smirk}, Deep3DFaceRecon \cite{deep3dfacerecon} and FOCUS \cite{focus}.

\vspace{-8pt}
\paragraph{NoW Evaluation.} Our results on NoW test set are presented in Table \ref{tab:nowresults}. \OURS{} outperforms all other methods that rely on only 2D supervision to train their head geometry predictor. Other methods, like TokenFace, show impressive performance, but these rely on using 3D scan data for direct supervision. \OURS{}'s performance shows promise in the direction of using more scalable 2D data for learning to predict geometry.

\vspace{-8pt}
\paragraph{Nersemble Evaluation.} The NoW dataset only assesses a method's ability to reconstruct a person's head geometry under a neutral expression. To evaluate each method's ability to reconstruct expressive head geometry, we evaluated them through our new Nersemble benchmark introduced in Section \ref{sec:Evaluation}. As shown in Table \ref{tab:nersemble_results}, \OURS{} outperforms all competing methods by a fairly substantial margin, highlighting its ability to predict accurate geometry even under varying poses and expressions.

% OLD DATA
% \begin{table}[h]
% \centering
% \begin{tabular}{@{}l|ccc@{}}
% \toprule
% Method & \multicolumn{3}{c}{Distance (mm)} \\ 
%        & Mean & Median & Std \\ 
% \midrule
% FOCUS \cite{focus} & xxx & xxx & xxx \\
% Deep3DFaceRecon \cite{deep3dfacerecon} & xxx & xxx & xxx \\
% DECA \cite{DECA} & 2.45 & 2.45 & 0.45 \\
% EMOCA v2 \cite{EMOCA} & 2.56 & 2.54 & 0.46 \\
% SMIRK \cite{smirk} & 2.26 & 2.21 & 0.41 \\
% Ours  & \textbf{2.06} & \textbf{1.99} & \textbf{0.39} \\
% \bottomrule
% \end{tabular}
% \caption{Distance metrics on our Nersemble-based~\cite{nersemble} reconstruction benchmark. Note that SMIRK~\cite{smirk} uses MICA~\cite{zielonka2022mica} for its shape code.}
% \label{tab:nowresults}
% \end{table}

% CROPPED
\begin{table}[t]
\centering
% \setlength{\tabcolsep}{18pt} % adjust to make table exactly same width as column
% \begin{tabular}{@{}l|ccc@{}}
\setlength{\tabcolsep}{13.5pt} % adjust to make table exactly same width as column
\begin{tabular}{*{4}{c}}
\toprule
\multirow{2}*{Method} & \multicolumn{3}{c}{Distance (mm)} \\ 
\cmidrule(lr){2-4}
       & Mean & Median & Std \\ 
\midrule
DECA \cite{DECA} & 2.46 & 2.46 & 0.46 \\
EMOCA \cite{EMOCA} & 2.58 & 2.54 & 0.46 \\
SMIRK \cite{smirk} & 2.30 & 2.24 & 0.42 \\
\textbf{Ours}  & \textbf{2.18} & \textbf{2.12} & \textbf{0.40} \\
\bottomrule
\end{tabular}
\vspace{-2mm}
\caption{Distance metrics on our Nersemble-based~\cite{nersemble} reconstruction benchmark. Note that SMIRK~\cite{smirk} uses MICA~\cite{zielonka2022mica} for its shape code.}
\label{tab:nersemble_results}
\end{table}

% NO CROP
% \begin{table}[h]
% \centering
% \begin{tabular}{@{}l|ccc@{}}
% \toprule
% Method & \multicolumn{3}{c}{Distance (mm)} \\ 
%        & Mean & Median & Std \\ 
% \midrule
% DECA \cite{DECA} & 2.64 & 2.60 & 0.48 \\
% EMOCA v2 \cite{EMOCA} & 2.73 & 2.68 & 0.47 \\
% SMIRK \cite{smirk} & 2.48 & 2.44 & 0.41 \\
% Ours  & \textbf{2.35} & \textbf{2.31} & \textbf{0.43} \\
% \bottomrule
% \end{tabular}
% \caption{Distance metrics on our Nersemble-based~\cite{nersemble} reconstruction benchmark. Note that SMIRK~\cite{smirk} uses MICA~\cite{zielonka2022mica} for its shape code.}
% \label{tab:nowresults}
% \end{table}

\vspace{-8pt}
\paragraph{AffectNet.} Finally, we evaluate the capacity of our model's estimated 3DMM parameters to predict emotion. Following EMOCA \cite{EMOCA}, and as described in Section \ref{sec:Evaluation}, we use \OURS{} and competing methods to predict 3DMM parameters for all of AffectNet. We then fit a 4-layer MLP to the training portion of these parameters, and evaluate on the test set. The results (Table \ref{tab:affectnet}) show that the 3DMM parameters predicted by \OURS{} are rich in emotional content, outperforming both SMIRK~\cite{smirk} and EMOCA~\cite{EMOCA} in emotion prediction on AffectNet.

\begin{table}[t]
\centering
\setlength{\tabcolsep}{4pt} % adjust to make table exactly same width as column
% \begin{tabular}{@{}l|c@{\hskip 5pt}c|c@{\hskip 5pt}c|c@{}}
% \multirow{2}*{Model} & \multicolumn{2}{c}{Valence} & \multicolumn{2}{c|}{Arousal} & Emo \\
%       & CCC↑ & RMSE↓ & CCC↑ & RMSE↓ & Acc. \\
\begin{tabular}{*{6}{c}}
\toprule
\multirow{2}*{Model} & \multicolumn{2}{c}{Valence} & \multicolumn{2}{c}{Arousal} & Emo \\
                        \cmidrule(lr){2-3} \cmidrule(lr){4-5}  \cmidrule(lr){6-6}
                    & CCC↑ & RMSE↓ & CCC↑ & RMSE↓ & Acc. \\
\midrule
SMIRK~\cite{smirk} & 0.56 & 0.29 & 0.68 & 0.31 & 0.65 \\
EMOCA~\cite{EMOCA} & 0.58 & 0.28 & 0.70 & 0.31 & 0.68 \\
\textbf{Ours} & \textbf{0.62} & \textbf{0.27} & \textbf{0.74} & \textbf{0.30} & \textbf{0.70} \\
\bottomrule
\end{tabular}
\vspace{-2mm}
\caption{Emotion recognition performance on AffectNet~\cite{mollahosseini2017affectnet}, we report the concordance correlation coefficient (CCC) and root mean squared error (RMSE) on predicting Valence and Arousal, plus accuracy in 8-way emotion classification.}
\label{tab:affectnet}
\vspace{-2mm}
\end{table}

\subsection{Ablation Study}

We perform an ablation over the most important hyperparameters of our method (Table \ref{tab:ablation}).
With only photometric loss and basic regularization, our method still outperforms DECA on the NoW validation set (median of 1.18mm \cite{DECA}), highlighting the effectiveness of Gaussians in comparison to differentiable mesh rendering for this type of task. Error rates can be greatly improved, however, by adding perceptual losses, most notably the identity perceptual loss. The impact of $\mathcal{L}_\textit{normals} + \mathcal{L}_\textit{depth}$ illustrates the importance of encouraging a tight coupling of the Gaussians geometry with that of the 3DMM.
Disabling all edges in our graph convolutional network (effectively turning it into an `MLP ResNet') also markedly worsened performance, showing the effectiveness of this specific architecture.
Finally the visual improvements thanks to densification and pruning, enabled by our novel architecture, add a final additional boost to performance.

\begin{table}[t]
\centering
% \begin{tabular}{@{}c|c|c|c|cc@{}}
% \toprule
% $\mathcal{L}_\textit{perc}$ & $\mathcal{L}_\textit{ID}$ & \makecell{Densification \\ and Pruning} & \makecell{$\mathcal{L}_\textit{normals}$ \\  $+ \mathcal{L}_\textit{depth}$} & \multicolumn{2}{c}{Distance (mm)} \\ 
%        &       &       &       & Median & Mean \\ 
\setlength{\tabcolsep}{1pt}
\begin{tabular}{*{7}{c}}
\toprule
\multirow{2}*{ $\mathcal{L}_\textit{perc}$ }
& \multirow{2}*{  $\mathcal{L}_\textit{ID}$ }
& Densify
& $\mathcal{L}_\textit{normals}$
& Graph
& \multicolumn{2}{c}{Distance (mm) } \\
\cmidrule(lr){6-7} % Adjusted cmidrule
&   &  /Prune  &   $+ \mathcal{L}_\textit{depth}$   & Enabled & Median & Mean \\
\midrule
\xmark & \xmark & \xmark & \xmark & \cmark & 1.09 & 1.38 \\
\cmark & \xmark & \xmark & \xmark & \cmark & 1.06 & 1.31 \\
\cmark & \xmark & \cmark & \cmark & \cmark & 1.03 & 1.25 \\
\cmark & \cmark & \xmark & \xmark & \cmark & 0.99 & 1.22 \\
\cmark & \cmark & \cmark & \cmark & \xmark & 0.98 & 1.21 \\
\cmark & \cmark & \cmark & \xmark & \cmark & 0.98 & 1.21 \\
\cmark & \cmark & \xmark & \cmark & \cmark & 0.96 & 1.20 \\
\cmark & \cmark & \cmark & \cmark & \cmark & \textbf{0.93} & \textbf{1.16} \\
\bottomrule
\end{tabular}
\vspace{-2mm}
\caption{Ablation study results on the NoW Validation Set. The table shows active components with checkmarks and corresponding distances.}
\label{tab:ablation}
\vspace{-4mm}
\end{table}

\section{Conclusion}
We introduced \OURS{}, a novel method for 3D head reconstruction and animation, learned in a self-supervised fashion from only monocular videos. Thanks to its use of Gaussian Splatting approach, \OURS{} receives a more detailed learning signal from photometric losses, in turn enabling it to learn a more accurate head geometry predictor.

However, our method faces certain limitations. While our Gaussian-based representation offers stronger photometric supervision compared to meshes, attaining perfect alignment of head and mesh in a feedforward manner remains challenging. This may limit the photorealism of the predicted Gaussians, as some blurring is required to account for any misalignment. Additionally, the lack of 3D supervision results in scale-free meshes, and the fixed-FOV assumption forces our model to use the predicted head shape to model head distortions owing to FOV differences. Future work could explore incorporating offline refinement of the predicted 3DMM meshes before rendering the rigged Gaussians. This may address shortcomings of feedforward prediction and lead to sharper renderings. Training with multiview data or direct 3D supervision is another promising direction. Finally, our method might benefit from an accurate image-based FOV predictor, like in \cite{CameraHMR}.

Overall, \OURS{} represents a significant step forward in the field of learning self-supervised 3D head reconstruction and animation from 2D data, providing a foundation for future advances and applications.

% but also sets new benchmarks in expressive head geometry reconstruction and emotional content evaluation.

% Working title

% BADG3R: Binding and Driving Gaussians with a 3DMM for Reconstruction of Human Heads (or binding and animating 2D gaussians)

% L3DGER: Learning 3D Geometry and Expression Reconstruction via 2D Gaussians

% L3DGER: Learning 3D Geometry and Expression Reconstruction via 2D Gaussians

% GRAP3FRUIt: Gaussian-dRiven Accurate Predictor for 3D Facial ReconstrUctIon

% Self-Supervised 2D Gaussians for Expressive 3D Reconstruction

% Binding and Animating Self-Supervised 

{\small
\bibliographystyle{ieeenat_fullname}
\bibliography{bibliography}
}

\end{document}